\documentclass[graybox]{SNmult}

\usepackage{type1cm}        % activate if the above 3 fonts are
\usepackage{makeidx}         % allows index generation
\usepackage{graphicx}        % standard LaTeX graphics tool
\usepackage{multicol}        % used for the two-column index
\usepackage[bottom]{footmisc}% places footnotes at page bottom

\usepackage{newtxtext}       % 
\usepackage[varvw]{newtxmath}       % selects Times Roman as basic font

\makeindex             % used for the subject index
\begin{document}

\title*{Your Robot Will Feel You Now: Empathy in Robots and Embodied Agents}
\author{Angelica Lim and Ö. Nilay Yalçin}
\institute{Angelica Lim \at School of Computing Science, Simon Fraser University, \email{angelica@sfu.ca}
\and Nilay Yalçin \at School of Interactive Arts and Technology, Simon Fraser University, \email{oyalcin@sfu.ca}}
\motto{Accepted for publication in ``Empathy and Artificial Intelligence: Challenges, Advances and Ethical Considerations" edited by Anat Perry; C. Daryl Cameron.}

\maketitle
\abstract*{The fields of human-robot interaction (HRI) and embodied conversational agents (ECAs) have long studied how empathy could be implemented in machines. One of the major drivers has been the goal of giving multimodal social and emotional intelligence to these artificially intelligent agents, which interact with people through facial expressions, body, gesture, and speech. What empathic behaviors and models have these fields implemented by mimicking human and animal behavior? In what ways have they explored creating machine-specific analogies? This chapter aims to review the knowledge from these studies, towards applying the lessons learned to today's ubiquitous, language-based agents such as ChatGPT.
}

\abstract{The fields of human-robot interaction (HRI) and embodied conversational agents (ECAs) have long studied how empathy could be implemented in machines. One of the major drivers has been the goal of giving multimodal social and emotional intelligence to these artificially intelligent agents, which interact with people through facial expressions, body, gesture, and speech. What empathic behaviors and models have these fields implemented by mimicking human and animal behavior? In what ways have they explored creating machine-specific analogies? This chapter aims to review the knowledge from these studies, towards applying the lessons learned to today's ubiquitous, language-based agents such as ChatGPT.
}

Computational empathy refers to artificial systems able to simulate empathic responses by recognizing and responding to emotional states. Empathy, in this context, can be understood as both the capacity to share another’s emotions and the ability to adopt their perspective—abilities that enable individuals to quickly relate to others, fostering cooperation and support in social contexts (de Waal \& Preston, 2017; Preston, 2007). Applications of computational empathy range from virtual counseling in healthcare to adaptive tutors in education (Paiva et al., 2017). In this chapter, we aim to provide a brief introduction to embodied computational empathy research in place since the early 1990s, specifically in the fields of human-robot interaction (HRI) and embodied conversational agents (ECAs) that interact with people through facial expressions, body, gesture, and speech. These artificial intelligence (AI) systems, which we define as agents capable of perceiving their environment and making decisions to act upon it (Russell \& Norvig, 2009), aimed to autonomously interact with humans. By modeling their empathic behaviors through embodiment, researchers have sought to improve not only the social compatibility of interactive agents but also our understanding of empathy mechanisms and links to physical grounding.

\section{Embodied Empathic Agents}
\label{sec:1}
Recent advances in large language models have spurred renewed interest in computational empathy. However, natural language interaction was an unsolved problem for many decades, and early embodied empathy research focused primarily on non-linguistic expressive behaviors accompanied with minimal linguistic interaction using theory-based, top-down approaches. Implementation of such behaviors depended on the agent’s embodiment, whether physical (in the case of robots) or virtual (in the case of software agents, including avatars or non-player characters), and which modalities are implemented for the interaction.

In the field of ECAs, a seminal work was the Sensitive Artificial Listener (SAL) developed under the SEMAINE project (Douglas-Cowie et al., 2008; Schroder et al., 2012). SAL was one of the first real-time, multimodal conversational systems that processed user input (i.e., video, audio, and speech) and provided facial and verbal feedback behaviors. The results showed that the full, multimodal system outperformed the control in terms of user preference, engagement, and appropriateness compared to the expressionless voices and faces. Many other studies showed that users expected ECAs to show empathic facial expressions during interaction (Niewiadomski et al., 2008) and respond with appropriate non-verbal emotion cues (Boukricha et al., 2013).

Lisetti and colleagues (2013) developed a virtual counselor designed to promote health behavior change, which adapts its verbal and nonverbal behaviors based on user inputs. In a controlled study, participants strongly preferred the empathic embodied agent over a neutral or text-based interface, particularly in terms of motivation, enjoyment, safety, and usability. However, embodied empathy must align with context to be effective. Users in Becker-Asano’s study on the game-playing agent MAX (Becker-Asano, 2008) found MAX’s “positive empathy” (cheering when the human won) unusual in a competitive setting. Interestingly, users were most irritated by a non-emotional agent, and their negative emotions decreased when MAX performed a calming gesture in response to frustration. M-Path, another real-time multimodal ECA, demonstrated that facial expressions alone were sufficient for low-level (affective) empathy, while linguistic expression did not significantly enhance perception of affective empathy (Yalçın \& DiPaola, 2019). 

In robotics, the canonical emotion-aware system was Kismet (Breazeal et al., 1990). Albeit lacking linguistic responses, the robot produced affectively appropriate responses in terms of bodily stance, vocal affect, and facial expression, inspiring subsequent social robotics work. Studies with empathic robots like iCat showed that users preferred robots displaying empathic nonverbal behavior (Cramer, 2010), which could increase interaction length with improved social presence, engagement, help, and self-validation (Leite et al., 2014). As empathic ECA and robot research advances (see a review in Paiva et al., 2017), the findings suggest that care should be taken when developing empathetic agents, as not only “what you say” but “how you say it” is important.

\begin{figure}[b]
%\sidecaption
% For example, with the graphicx style use
\includegraphics[width=\textwidth]{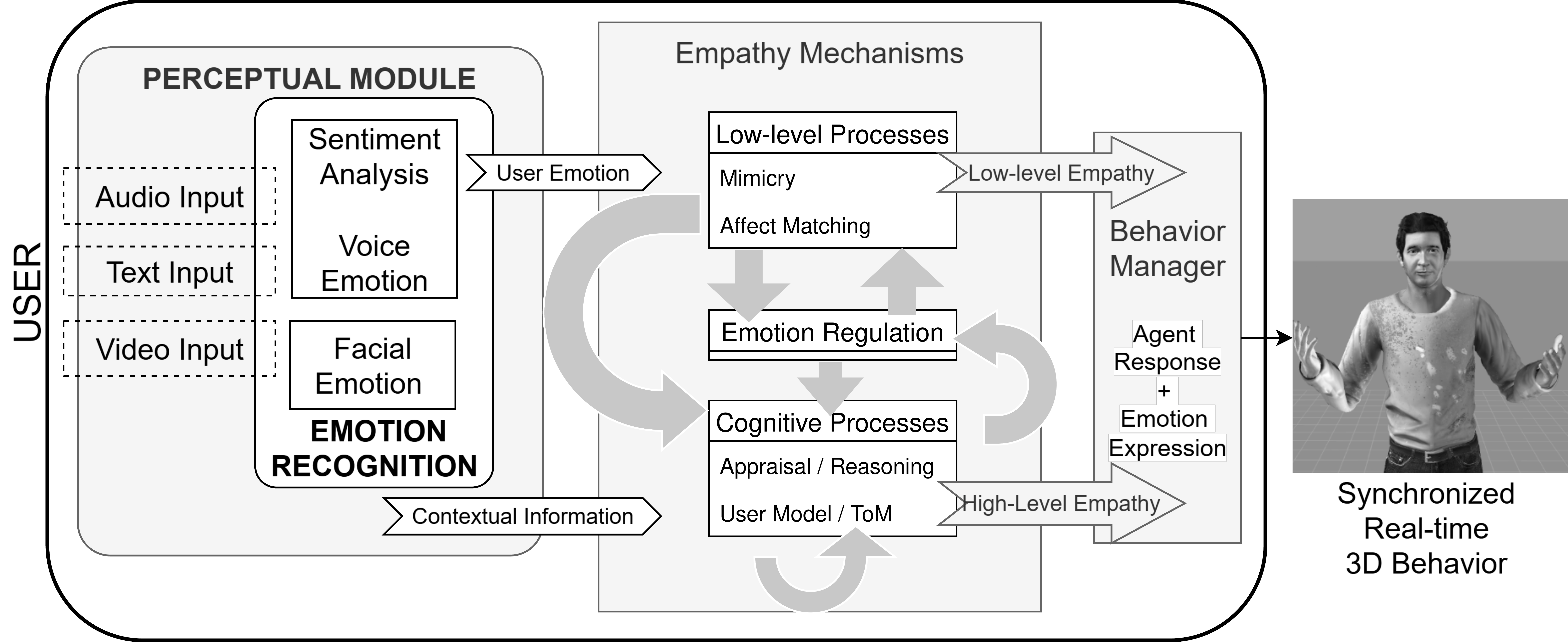}
\Description{This is figure Alt-Text for Figure 1.}
% Use the \Description command to include "alternative text" to describe the figure for readers with disabilites
\caption{An example of empathic agent architecture that incorporates a top-down approach to modeling low-level (affective) and high-level (cognitive) empathy.}
\label{fig:1}       % Give a unique label
\end{figure}

\subsection{Models of Computational Empathy}
\label{subsec:1}
Most prior research on top-down models of computational empathy was inspired from the theoretical understanding of how empathic behavior develops and functions in humans. Drawing from various fields including neuroscience, psychology, and ethology, empathy is often categorized into two primary types: affective empathy, which involves the automatic and unconscious mirroring or matching of others' emotions, and cognitive empathy, which requires an understanding of another’s emotional and mental state using mechanisms like perspective-taking and theory of mind (Coplan, 2011; De Waal \& Preston, 2017). However, creating systems capable of addressing both the affective and cognitive aspects of empathy remains a significant challenge. 

Early work aimed to create internal architectures representing the emotional logic of such systems. For instance, much work has focused on appraisal models, and we refer the reader to a recent review (Ojha, 2021) on computational emotion appraisal models such as OCC, CogAff and WASABI. Another approach was derived from the perception-action model (PAM) of empathy (Preston \& deWaal, 2002; Preston, 2007), where empathic behavior in embodied agents is modeled through their ability to perceive and express affective information (Yalcin \& DiPaola, 2020). Perception refers to an agent’s capacity to detect human emotions or their causes using various input modalities (e.g., visual, auditory, or tactile data). Expression, on the other hand, is how the agent communicates empathy. However, the key to modeling different levels of empathy lies in the processing that maps perception to expression, rather than expression alone. Fig. 1 shows an example PAM-based empathic agent architecture with the perceptual and expression capabilities of an embodied agent, where a behavior controller decides the agent behavior given inputs. In recent years, end-to-end models using Deep Neural Networks (DNNs) have also been used to bypass a need for a behavior controller, and directly tie the perception of the agent to its behaviors, often using only text-based interaction.

\subsection{Current Challenges}
\label{subsec:2}
With recent advances in Large Language Models (LLMs), there has been a major shift in the field. LLMs trained on large-scale datasets now demonstrate impressive linguistic empathy, producing context-aware responses that mimic human emotional understanding. Unlike top-down theory driven approaches, modern LLMs use a bottom-up approach that generates end-to-end empathic dialogue by learning from vast amounts of natural language data.  However, these models often do not have the control and theoretical grounding of the earlier top-down embodied empathy research. In addition, real-time, synchronous behavior in embodied systems remains a significant challenge, especially in robotics, where processing delays, hardware limitations, and integration difficulties hinder seamless interaction.

Both approaches focus on how the emotions and empathy are expressed, rather than the genuine feeling of the emotions. However, one of the defining characteristics of human empathy is that the “feeling” of the observer is congruent with the emotion and state of the object of empathy(Cuff et. al., 2016). This raises an important question in computational empathy: can artificial systems truly feel emotions, or are they only simulating behavior? This mirrors debates in Cognitive Science about the distinction between genuine understanding and mere appearance, as seen in Searle's Chinese Room Argument (Searle, 1980; Cole, 2024), which suggests a machine may appear to understand language but lacks true comprehension. Similarly, in computational empathy, the question arises: if an agent produces empathetic responses, is it truly ``feeling" empathy, or just processing input based on rules or models?

\section{Can Embodied agents truly understand, feel or care?}

One of the historically pernicious issues of embodied artificial empathy has been whether these agents could truly empathize with humans. For example, if a robot says, “I'm sorry you had a bad day,” the human could ask, “Do you really feel sorry?” Among humans, showing fake emotion can be considered manipulative, and theorist Sherry Turkle has strongly opposed machines that show an emotional façade. To address the challenge of fake empathy, then, developing “authentic” artificial empathy suffers from at least two major concerns. First, could we implement it, and if so, how? And secondly, should we?

How could we implement authentically empathetic robots?
Previously, Lim argued that authenticity requires not only reproducing the essential features of an object but also preserving original methods used in its creation (Lim and Okuno, 2015). For instance, a theme park castle might successfully replicate the defining features of a castle—such as towers and parapets. However, its authenticity could be more convincing if skilled artisans employed techniques similar to those used by 18th-century builders. Likewise, the perception of authentic empathy may depend on both reproducing its essential characteristics and mirroring the processes through which genuine empathy arises.

\subsection{Essential features of embodied empathy}

What are the essential features of feeling sorry? In the previous section, we discussed the physically perceptible changes in the previous section. Now, we turn to the most challenging and elusive aspect: the feeling itself. Later, we will examine what we know of neural elements that connect these feelings to the external world.

\textbf{Feeling.} Could a robot ever feel? What does feeling mean, and can we define it for humans? We acknowledge that these questions point to a larger literature dubbed as the “hard problem” of consciousness that includes “the felt quality of emotion” (Chalmers, 2007). Some neuroscientists such as Damasio, believe that the hard problem obstructs progress by making the issue appear impossible to solve\footnote{  https://nautil.us/whats-so-hard-about-understanding-consciousness-238421/}. According to Damasio, feelings are “the expression of human flourishing or human distress, as they occur in mind and body.” (Damasio, 2003). Notably, Damasio emphasizes the body, suggesting that embodiment—a physical presence—is essential for experiencing feelings. If embodiment is essential for feelings, as Damasio suggests, it provides a useful lens for exploring how robots might experience analogous states.

A natural starting point is distress, which can be understood through the lens of homeostasis. For instance, newborn infants cry when their physiological balance is threatened; they may be hungry, too cold, too hot, or uncomfortable from a wet diaper. This cry signals distress, reflecting a state where homeostasis and survival are at risk. Similarly, Lim proposed that a robot could experience a comparable state of physical distress. A low battery might be analogous to hunger, while elevated motor temperatures could be likened to feeling too hot or too cold. In both cases, deviations from optimal operating conditions threaten the robot’s ability to function, paralleling Damasio’s definition of “distress as it occurs in the body.”

Many robots, such as a Roomba returning to its charging station, already handle such distressed states by restoring homeostasis, i.e. returning to a “flourishing” state. Yet, despite this behavior, we do not readily attribute feelings to a vacuum robot with a low battery. This raises a fundamental question: what distinguishes a robot experiencing low battery from a robot feeling bad about having a low battery?

\textbf{Neural elements}. Beyond bodily feeling, Damasio suggests that feelings are also expressions of human flourishing or distress as they occur in the mind. The mind evaluates sensory signals reaching the brain, interpreting distressing signals as negative. In humans, a key brain region involved in this process is the insula. Damasio and other affective neuroscientists propose that the insula maps visceral states to emotional experiences, and is a key component of conscious feelings. This region monitors the body’s condition, including visceral states and interoceptive signals such as heat, pain, and muscle ache. The insula is also implicated in feeling pain, experiencing parental love, and empathizing with others. Moreover, it is thought to convert unpleasant tastes or smells into disgust and transform sensual touch into pleasure. Thus, the insular cortex may function as a neural appraiser, interpreting bodily states of “distress” or “homeostasis” as “negative” or “positive,” and conveying this information for downstream processing.

\textbf{How could an insula be implemented in a robot?} A basic artificial insular cortex could serve as a central hub, integrating valenced associations between external stimuli, internal states, and bodily conditions. Consider a robot that signals low battery with a blinking red light: rather than directly linking the battery state to the light, this process could be routed through a central “insula.” Similarly, if the robot becomes stuck, this state could be processed through the insula and trigger the same red light. Akin to animal expressions of feeling, these expressive behaviors could also be adaptive. For instance, a Roomba might reduce its maximum velocity to extend its functional lifespan when the battery is low, much like a dog with low energy lying down to conserve energy.

Yet, despite these parallels, pre-programming a lookup table of aversive states and corresponding behaviors presumably still falls short of creating a robot that truly feels. As with the theme park castle, simply “copying” a reactive system from one entity to another does not seem sufficient to achieve genuine authenticity.

\textbf{Development of empathy.}
We argued that authenticity also requires a creation process that mirrors the original. One way to approach this challenge is by examining infant emotional development. In experiments with the Leonardo robot, Cynthia Breazeal’s group employed a developmental robotics paradigm to teach the robot to associate objects with positive and negative outcomes. Using a somatic marker paradigm and infant-directed speech, the robot learned to associate aversive stimuli, such as a “bad” Cookie Monster, and positive stimuli, such as a “good” Big Bird (Breazeal, 2005). This example, along with Minoru Asada’s work on developing artificial empathy in robots (Asada, 2015), provides a foundation for how insights from developmental psychology can inform the creation of computational empathy built through experience.

In developmental robotics, two key components are the learning apparatus itself (i.e., the brain) and the social interactions that support learning. Studies have shown that human emotions develop through interactions such as infant-directed speech, mirroring, and labeling. Similarly, a system that learns through social interaction—connecting expressions and concepts grounded in its social environment to its embodied, low-level physical “feelings”—may be perceived as having a deeper understanding and capacity for feeling. We implemented a basic version of a “feeling” empathetic robot (Lim and Okuno, 2015), with mirroring, an artificial insula, physical states and a developmental process with human caregivers expressing praise, prohibition, comfort and attention-seeking. Yet, more complete emotional development in robots, much like in humans, requires extensive learning from the ground up, scaffolding from low-level feelings to more complex expressions and concepts of emotion.

The requirement for actual experience addresses the question: "How could you understand how I feel if you’ve never felt it yourself?" This touches on the concept of sympathy, or feeling sorry for someone without having experienced the same situation. Perhaps a robot that has lived through experiences, encountering situations that evoke physical or emotional responses, would be more authentic than a robot or system that does not. Sci-fi depictions of robots with plausible emotional systems often feature robots that have real-life interactions and experiences with humans. However, unlike humans, robots have the ability to build upon each other’s learning through mechanisms like collective learning and fine-tuning from checkpoints.

\subsection{Should we develop authentically empathetic robots?}

In this chapter, we first defined empathy from a functional perspective, and this working definition lends itself well to building practical, useful computational empathic AI systems. However, it can be argued that for robots to authentically ``feel sorry," they must possess the capacity to actually feel. This could entail the need for life experience—whether their own or as part of a ``collectively shared" AI experience—interacting with humans to learn and experience distress. However, two major issues arise when asking the question, ``Should we develop authentically empathetic robots?”. First, this approach would be akin to creating robots capable of feeling pain, which is morally fraught. Would this not be similar to inflicting pain on them? Is it necessary? From a consequentialist perspective, does the benefit justify the potential harm? Second, such a development could lead to the creation of an entity driven to reduce its own pain, essentially resulting in an AI agent with a survival instinct. This could create agents motivated to eliminate the sources of their pain, potentially including the humans who seek empathy from these AI systems in the first place.
In sum, the field of human-centered AI has long aimed to develop agents that support and empower humans rather than replace them. While embodied empathic agents that emulate humans through anthropomorphic features are undeniably compelling, engaging, and effective, we must carefully consider the costs and benefits of creating such systems to ensure they serve humanity in meaningful and ethical ways.

%%%%%%%%%%%%%%%%%%%%%%%% referenc.tex %%%%%%%%%%%%%%%%%%%%%%%%%%%%%%
% sample references

\end{document}